\documentclass[letterpaper, 10 pt, conference]{ieeeconf}  % Comment this line out if you need a4paper

\IEEEoverridecommandlockouts                              % This command is only needed if 
\usepackage{booktabs}
\usepackage{graphics} % for pdf, bitmapped graphics files
\usepackage{graphicx}

\usepackage{epsfig} % for postscript graphics files
\usepackage{times} % assumes new font selection scheme installed
\usepackage{amsmath} % assumes amsmath package installed
\usepackage{bm}  % for bold math symbols
\usepackage{amssymb}  % assumes amsmath package installed
\usepackage{arydshln} % Dashed Mid rule
\usepackage{mathrsfs}
\usepackage{eucal}
\usepackage{capt-of}
\usepackage[dvipsnames]{xcolor}
\usepackage{xcolor,colortbl}
\usepackage[accsupp]{axessibility}  % Improves PDF readability for those with disabilities.
\usepackage{cite}
\usepackage{float}
\makeatletter
\let\NAT@parse\undefined
\makeatother
\definecolor{citepurple}{rgb}{0.288,0.1196,0.7}
\usepackage[backref=page,breaklinks,colorlinks,bookmarks,citecolor=citepurple]{hyperref}
\usepackage{booktabs,arydshln}
\makeatletter
\def\adl@drawiv#1#2#3{%
        \hskip.5\tabcolsep
        \xleaders#3{#2.5\@tempdimb #1{1}#2.5\@tempdimb}%
                #2\z@ plus1fil minus1fil\relax
        \hskip.5\tabcolsep}
\newcommand{\cdashlinelr}[1]{%
  \noalign{\vskip\aboverulesep
           \global\let\@dashdrawstore\adl@draw
           \global\let\adl@draw\adl@drawiv}
  \cdashline{#1}
  \noalign{\global\let\adl@draw\@dashdrawstore
           \vskip\belowrulesep}}
\makeatother

\usepackage{boldline}  % bold lines in tables
\usepackage{multirow}
\definecolor{Gray}{gray}{0.90}
\newcolumntype{g}{>{\columncolor{Gray}}c}
\definecolor{ffe1da}{RGB}{255,225,218}
\definecolor{F7E0D5}{RGB}{247,224,213}
\definecolor{darkF7E0D5}{RGB}{209,154,128}
\colorlet{Light}{White!0!F7E0D5}

\newcommand{\symbolHt}{1.0em}
\newcommand{\SeasonChar}{%
  \begingroup\normalfont
  \includegraphics[height=\symbolHt]{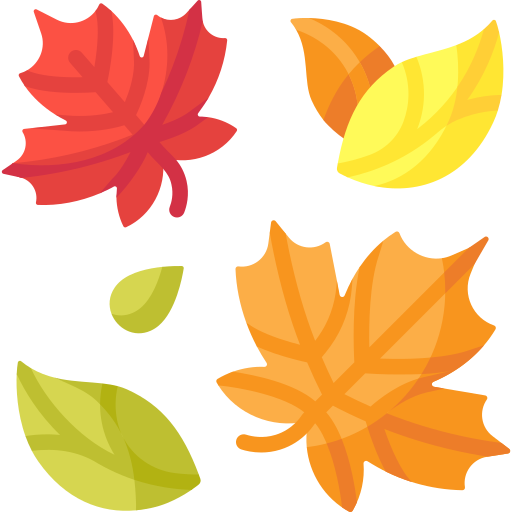}%
  \endgroup
}
\newcommand{\LightChar}{%
  \begingroup\normalfont
  \includegraphics[height=\symbolHt]{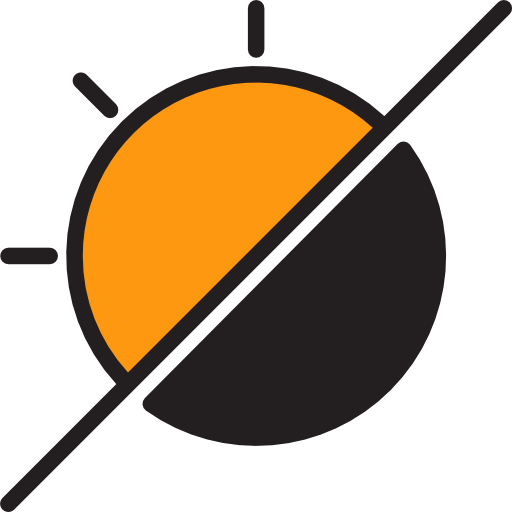}%
  \endgroup
}
\newcommand{\AltitudeChar}{%
  \begingroup\normalfont
  \includegraphics[height=\symbolHt]{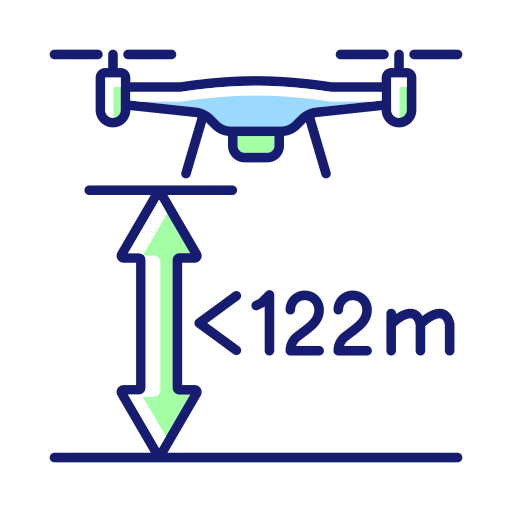}%
  \endgroup
}
\newcommand{\TrajChar}{%
  \begingroup\normalfont
  \includegraphics[height=\symbolHt]{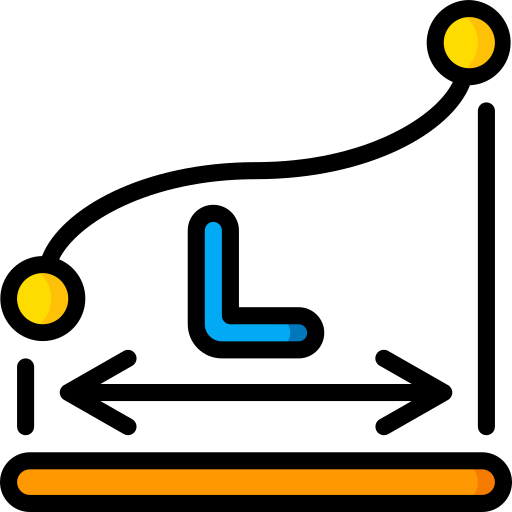}%
  \endgroup
}
\newcommand{\WeatherChar}{%
  \begingroup\normalfont
  \includegraphics[height=\symbolHt]{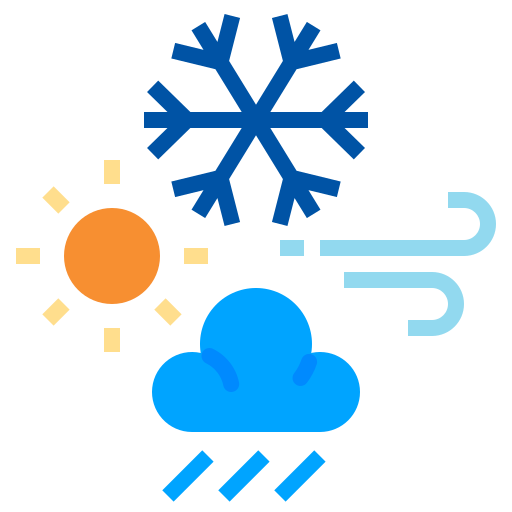}%
  \endgroup
}

\newcommand{\AllChar}{%
  \begingroup\normalfont
  \includegraphics[height=\symbolHt]{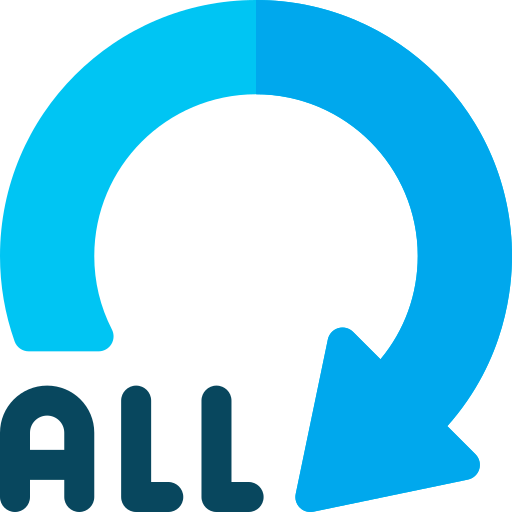}%
  \endgroup
}

\definecolor{darkpurple}{rgb}{0.288,0.1196,0.7}

\definecolor{amber}{rgb}{1.0, 0.75, 0.0}

\usepackage[capitalize]{cleveref}
\crefname{section}{Section}{Sections}
\crefname{table}{Table}{Tables}

\makeatletter
\newcommand{\cdashmidrule}[1]{%
  \noalign{\vskip\aboverulesep}
  \cdashline{#1}
  \noalign{\vskip\belowrulesep}}
\makeatother

\newcommand*{\subfigref}[2][]{%
  Fig. \hyperref[{fig:#2}]{%
    \ref*{fig:#2}%
    \ifx\\#1\\%
    \else
      \,#1%
    \fi
  }%
}

\newcommand{\coolname}{\textit{FoundLoc}}

\graphicspath{{figures/}} 

\newcommand{\authorhref}[3][citepurple]{\href{#2}{\color{#1}{#3}}}

\title{\LARGE \bf
\coolname{}: Vision-based Onboard Aerial Localization in the Wild
}

\author{
\authorhref{https://shockwaveHe.github.io/}{Yao He$^{*}$}, 
\authorhref{https://www.ivancisneros.com/}{Ivan Cisneros$^{*}$},
\authorhref{https://nik-v9.github.io/}{Nikhil Keetha},
\authorhref{https://www.jaypatrikar.me/}{Jay Patrikar},
\authorhref{https://www.linkedin.com/in/zelinye}{Zelin Ye},
\\
\authorhref{https://www.linkedin.com/in/ian-higgins-53957718a}{Ian Higgins},
\authorhref{http://www.huyaoyu.com/}{Yaoyu Hu},
\authorhref{https://parvkpr.github.io/}{Parv Kapoor}, and
\authorhref{https://theairlab.org/}{Sebastian Scherer}
\\[5 pt]
Carnegie Mellon University
\thanks{$^*$Equal Contribution}% <-this % stops a space
}

\begin{document}

\makeatletter
\let\@oldmaketitle\@maketitle
\renewcommand{\@maketitle}{\@oldmaketitle
\centering
\begin{tabular}{cccc}
\includegraphics[trim={0cm 0cm 0cm 0cm},clip,width=\linewidth]{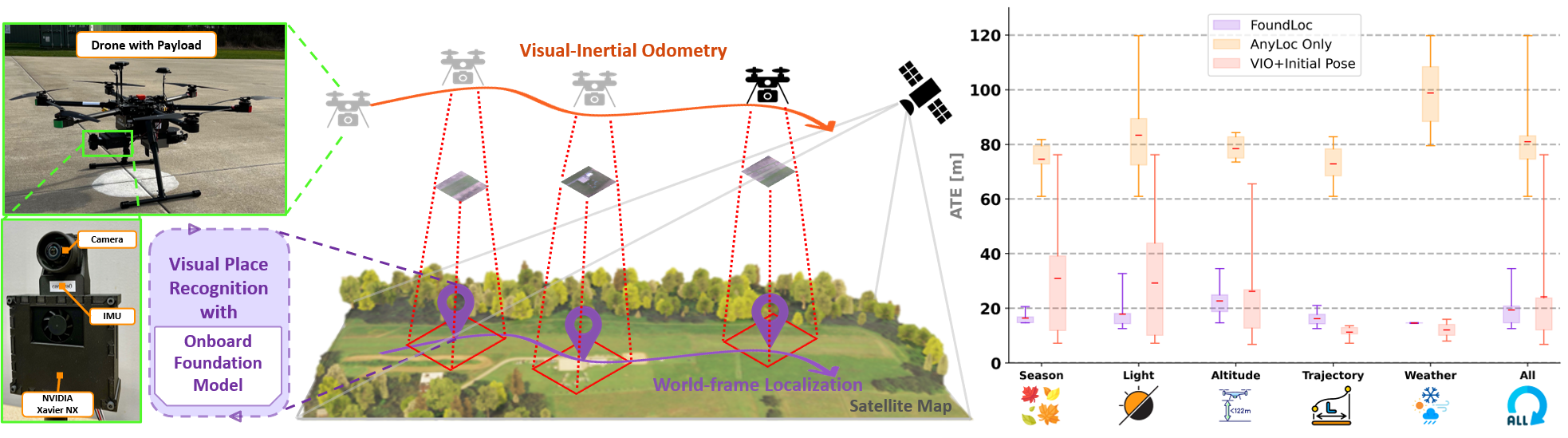}
\end{tabular}
\captionof{figure}{\textbf{\coolname{}} enables Unmanned Aerial Vehicle global localization in the wild using only a low-cost onboard vision-based system without relying on external GNSS signals. It achieves GNSS-denied localization by anchoring a Visual-Inertial Odometry trajectory into the world frame using a Visual Place Recognition module and a satellite map (\textit{middle}). Notably, \coolname{} is powered by a \textit{foundation model} to tackle this challenging \textit{Kidnapped Robot Problem}. In this work, we deploy our pipeline onto a custom compute-limited hardware payload and demonstrate it running in real-time (\textit{left}). As shown in the box plot (\textit{right}), our experiments demonstrate \coolname{}'s capability of dealing with the visual challenges arising from different seasons, lighting conditions, altitudes, trajectory patterns, and weather.
}
\label{fig:splash}
}
\makeatother

\maketitle
\thispagestyle{empty}
\pagestyle{empty}

%%%%%%%%%%%%%%%%%%%%%%%%%%%%%%%%%%%%%%%%%%%%%%%%%%%%%%%%%%%%%%%%%%%%%%%%%%%%%%%%

\begin{abstract}
Robust and accurate localization for Unmanned Aerial Vehicles (UAVs) is an essential capability to achieve autonomous, long-range flights. Current methods either rely heavily on GNSS, face limitations in visual-based localization due to appearance variances and stylistic dissimilarities between camera and reference imagery, or operate under the assumption of a known initial pose. In this paper, we developed a GNSS-denied localization approach for UAVs that harnesses both Visual-Inertial Odometry (VIO) and Visual Place Recognition (VPR) using a foundation model. This paper presents a novel vision-based pipeline that works exclusively with a nadir-facing camera, an Inertial Measurement Unit (IMU), and pre-existing satellite imagery for robust, accurate localization in varied environments and conditions. Our system demonstrated average localization accuracy within a $20$-meter range, with a minimum error below $1$ meter, under real-world conditions marked by drastic changes in environmental appearance and with no assumption of the vehicle's initial pose. The method is proven to be effective and robust, addressing the crucial need for reliable UAV localization in GNSS-denied environments, while also being computationally efficient enough to be deployed on resource-constrained platforms.

\end{abstract}

\setcounter{figure}{1} % Hotfix, for consistent figure numbers (else, after the teaser fig, latex skips to fig 3)

\graphicspath{{figures/}} 
\section{Introduction}

% Context
Unmanned aerial vehicles (UAVs) have increasingly become integral in a variety of applications, ranging from agriculture to emergency response. A foundational element enabling their versatility is accurate localization, typically provided by Global Navigation Satellite Systems (GNSS). However, GNSS solutions are not without their limitations; they are vulnerable to jamming, spoofing, and environmental interference that obstruct radio signals. Therefore, the ability of UAVs to infer their location in GNSS-denied situations is essential to creating reliable and fully autonomous systems. 

% Need
Vision-based methods for UAV GNSS-denied localization are a promising solution because cameras, being passive sensors, do not suffer from the same drawbacks as GNSS-based systems, while also being low SWaP-C (Size, Weight, Power, and Cost). Extensive research in visual-inertial odometry (VIO) and Simultaneous Localization and Mapping (SLAM) \cite{mourikis2007multi, davison2007monoslam, bloesch2015robust, jones2011visual, hesch2014camera, geneva2020openvins, qin2018vins, campos2021orb, rosinol2020kimera, forster2014svo, von2018direct} has yielded compelling evidence regarding the capacity of robots to achieve self-localization using only cameras and inertial measurement units (IMUs) in GNSS-denied environments. However, SLAM cannot provide Earth-fixed coordinates without external georeferencing. The accumulation of odometry drift is also a prominent concern as the robot's trajectory extends a long distance.  One promising vision-based method for aerial GNSS-denied localization is Visual Terrain-Relative Navigation (VTRN) \cite{fragoso2021seasonally, patel2020visual, choi2020brm, kinnari2022season, jurevivcius2019robust, kinnari2021gnss, shan2015google}. However, most VTRN methods assume the availability of initial position and heading information, as well as minimal odometric drift. This assumption permits pose propagation, and image registration or similarity estimation locally around the current position without resorting to a comprehensive global database search. Furthermore, challenges such as \textit{appearance variance}, \textit{stylistic dissimilarities}, and \textit{recurring visual patterns} between vehicle camera and satellite images lead to unreliable image registration and similarity estimation. There is a need for a vision-based solution that is generalizable enough to address the above problems.

% Task + Objective
In this paper, we investigate and implement a GNSS-denied localization pipeline that relies on VIO and Visual Place Recognition (VPR) with the generalizability of Foundation Models~\cite{Bommasani2021-go}. VPR has shown the capability of providing georeferenced image matches under large visual and viewpoint differences \cite{keetha2023anyloc}. Unlike image registration and similarity estimation that rely on local feature retrieval and matching, VPR works with aggregated global features which are better for capturing the high-level information of an image. However, traditional learning-based VPR methods, which are usually trained on limited datasets, cannot guarantee accurate localization because of the stylistic diversity among reference satellite imagery. Furthermore, the stylistic dissimilarities between camera images and satellite images significantly reduce the matching accuracy.

Our approach, termed \textbf{\textit{FoundLoc}}, is a GNSS-denied localization pipeline that attempts to tackle the aforementioned problems. The main contributions of this work are:
\begin{itemize}
    \item We propose a novel vision-based approach to achieve GNSS-denied localization for UAVs using pre-existing satellite imagery. We design our pipeline with an In-the-Wild assumption, i.e., the vehicle does not know its initial position, and the only sensors and knowledge it has are camera images, IMU readings, and a preloaded georeferenced satellite image database.

    \item We formulate a Selective Ordered Top-$N$ Recall evaluation metric, that takes into consideration whether a sufficient number of ground truth matches are present in the top $N$ of the VPR retrieval sequence. This scoring metric helps to better evaluate VPR performance for the purposes of localization.

    \item We conduct extensive real-world experiments to demonstrate the robustness of our algorithm to effectively operate under In-the-Wild scenarios.  With our experiments, we demonstrate the application of a foundation model in robotics tasks on computationally limited hardware. We also release our dataset, which we term ``\textit{Nardo-Air}'', that was used for these experiments.
\end{itemize}

% Conclusion
By leveraging the strengths of a highly generalizable VPR module, our \textit{FoundLoc} pipeline aims to address the limitations and challenges inherent in existing GNSS-denied, vision-based localization methods for UAVs, thereby offering a more robust and adaptable solution for real-world applications.

\section{Related Work}
\label{Sec:Related Work}
% Define Visual Terrain Relative Navigation abbreviation in preceding overview
\subsection{Visual-Inertial-Odometry (VIO) \& VI-SLAM}

Recent advancements in VIO or VI-SLAM have demonstrated remarkable performance in robot state estimation in GNSS-Denied environments. These achievements primarily leverage only camera and IMU inputs. Broadly categorized, VIO can be classified into two frameworks: filter-based \cite{mourikis2007multi, davison2007monoslam, bloesch2015robust, jones2011visual, hesch2014camera, geneva2020openvins} and graph-based optimization frameworks \cite{qin2018vins, campos2021orb, rosinol2020kimera, forster2014svo, von2018direct}.

Filter-based approaches rely on the Extended Kalman Filter (EKF) and exhibit notable processing speed advantages over graph-based approaches. On the other hand, graph-based optimization approaches construct a factor graph of states within a given time window and perform a maximum likelihood estimation. For nonlinear problems, graph-based approaches are more accurate than filter-based methods but require more computational resources. Both of these approaches exhibit drift due to linearization in marginalization steps and outliers \cite{forster2016manifold}. Also, VIO constructs its frame with its initial state as world origin, which is independent of the Earth frame. Therefore, UAVs cannot use only VIO or VI-SLAM to obtain accurate Earth-fix coordinates.

\subsection{Visual Terran Relative Navigation (VTRN)}

Previous vision-based UAV localization methods are usually referred to as VTRN \cite{fragoso2021seasonally, patel2020visual, choi2020brm, kinnari2022season, jurevivcius2019robust, kinnari2021gnss, shan2015google}. Most VTRN methods adopt a filter-based pipeline, where odometry measurements provide information for pose updating. In pose correction, some approaches construct correctors through image registration between camera and satellite images. 
%For better image registration, some VTRN methods will learn common features between query and references.
\cite{fragoso2021seasonally} trains a Siamese U-Net using diverse seasonal imagery datasets so that it transforms images into a season-invariant domain and conducts image registration in this domain. \cite{patel2020visual} directly conducts image registration between a camera and a satellite image by minimizing the Normalized Information Distance (NID) of the two images. \cite{choi2020brm} constructs a building ratio map (BRM) that captures the geometric features of the building area. Instead of estimating the relative transformation through registration, \cite{choi2020brm} uses the global pose estimated by BRM as a correction term. 

In particle-filter-based methods, the particles are interpreted as weighted reference images or poses. \cite{kinnari2022season} learns season-invariant features to calculate the similarity scores of each reference image. The similarity scores are later converted to particle weights. \cite{jurevivcius2019robust} uses normalized cross-correlation metric (NCC) to calculate image similarity and converts the similarity to particles' likelihood. \cite{kinnari2021gnss} generates ortho-projection from camera images and uses cross-correlation-based methods for matching score estimation between ortho-projection and satellite images.  
\cite{shan2015google} estimates the Histograms of Oriented Gradients (HOG) of images and conducts a coarse to fine search for local particle weighting.

% There are other form of VTRN methods. \cite{goforth2019gps} extracts deep feature representation of images using a CNN then adds the deep feature error into the odometry optimization problem to jointly update UAVs' motion parameters. \cite{autoEncodeSatellite} design an network to encode images. It uses an inner-product kernel to estimate similarity score of each reference images and the weighted sum of reference positions is the localized position.  

Although these VTRN methods demonstrate promising localization accuracy within their designated testing scenarios, their applicability is constrained by certain limitations that hinder their generalization. Firstly, these methods usually exhibit hard assumptions on odometry to relax the problem: (1). The initial position is available within a small error range, (2). odometry heading is known. These assumptions allow the filter to directly propagate position without a global search procedure, and image registration or particle weighting can be conducted locally around the current location. However, in terms of localization, robots are dealing with a kidnap problem where there is no prior knowledge of the initial position and heading. 

On the other hand, several challenges arise due to the disparity between the nadir-facing camera images and the satellite images. These challenges can be categorized as \textit{appearance variances}, \textit{stylistic dissimilarities}, and \textit{recurring visual patterns}. We will define these terminologies in ~\cref{Subsec:VPR}. VTRN methods are insufficient in addressing these challenges, as they focus on local features that inadequately represent the entirety of the images.

\subsection{Visual Place Recognition (VPR)}

VPR is usually defined as an image retrieval problem where the context is to recognize previously seen places solely based on images. A basic VPR pipeline achieves image matching by computing image-wise descriptors and then calculating descriptor similarity between queries and references \cite{vprtut}. This pipeline offers the best trade-off between matching accuracy and search efficiency \cite{garg2021your}.

Recent VPR methods involve using a feature extraction backbone network followed by a trainable aggregation layer. One notable aggregation method is NetVLAD \cite{arandjelovic2016netvlad}, which is a learning-based variant of the Vector-of-Locally-Aggregated Descriptors (VLAD) \cite{arandjelovic2013all}, where local features are softly assigned to a learned set of clusters. Powered by deep learning and large-scale VPR-specific data, VPR achieves a substantial boost in performance. DeLF \cite{noh2017large} and DeLG \cite{cao2020unifying} achieve large-scale image retrieval after training on Google-Landmark V1 (1 million images) and V2 datasets \cite{weyand2020google} (5 million images). TransVPR \cite{wang2022transvpr} and R2former \cite{zhu2023r2former}, based on vision Transformer \cite{dosovitskiy2020image}, achieve significant improvements in urban and suburban environments with 1.6 million street images from MSLS datasets \cite{warburg2020mapillary}. Similarly, CosPlace \cite{berton2022rethinking} and MixVPR \cite{ali2023mixvpr} achieve SOTA performance after training on 40 million images and 530,000 images, respectively.

Scaling up VPR-specific training datasets has been shown to be effective in improving performance. However, the aforementioned methods are usually environment-specific and task-specific, limiting their generalization capability. Trained on large and diverse datasets with self-supervision, foundation models have shown the ability to produce generalizable solutions for individual machine learning problems \cite{bommasani2021opportunities}. Benefits from the emerging foundation models, \cite{keetha2023anyloc} proposes the first universal VPR solution AnyLoc that exhibits anywhere, anytime, and anyview capacities without any task or condition-specific training. In this paper, we adopt AnyLoc to achieve In-the-Wild satellite image recognition and provide geo-reference positions.

\graphicspath{{figures/}} 
%Yao
\begin{figure}[!ht]
\centering
\includegraphics[width=1\linewidth]{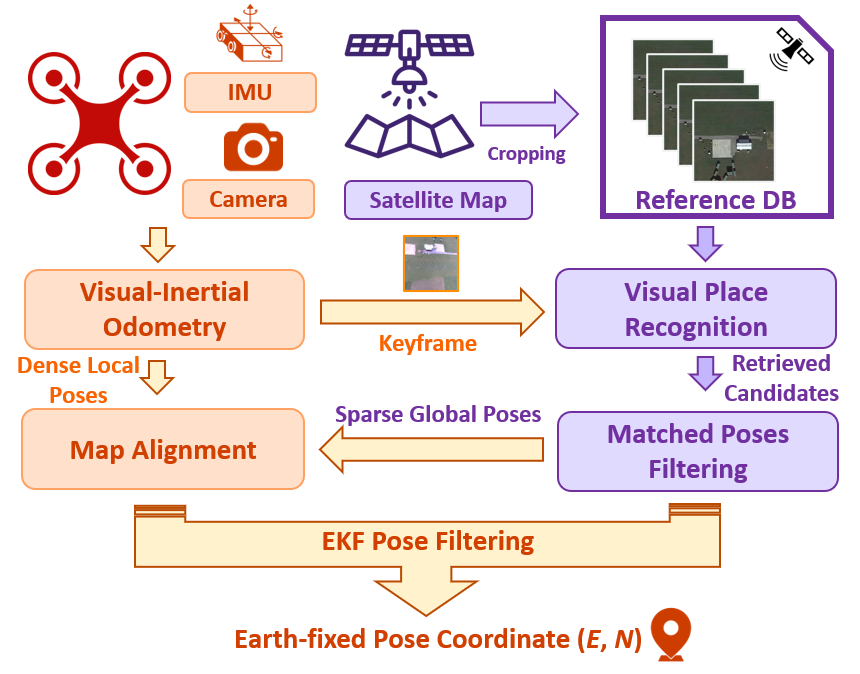}
\caption{\footnotesize The system diagram for our GNSS-Denied localization pipeline.}
\label{system}
\end{figure}

\section{Approach}
\label{Sec:approach}
In our formulation, we estimate the Earth-fixed coordinates $($Easting, Northing$)$ of a UAV's position using a camera, an IMU, and pre-loaded satellite imagery. We operate under the following assumptions: 
\begin{itemize}
    \item The UAV has no knowledge of its initial position and global heading.
    \item Images in the satellite imagery database only provide corresponding Earth-fixed coordinates. 
    \item The altitudes of reference images are unknown. 
    \item The camera parameters of satellite images are unknown. We do not assume a virtual camera for satellite maps because of the stylistic difference between real camera images and satellite images.
\end{itemize}

We denote the VIO body position in the odometry frame as $\boldsymbol{P}_i^{{L}} = (x_i^{L}, y_i^{L})^{T}$, where $i$ indicates the $i-$th camera query image and ${L}$ represents the local VIO frame. The global earth-fixed coordinate provided by each satellite image is denoted as $\boldsymbol{P}_k^{W} = (x_k^{W}, y_k^{W})$, where $k$ indicates the $k-$th reference satellite image and $W$ represents the Earth frame. $\mathcal{L}^q = \{\boldsymbol{I}_1^q,\boldsymbol{I}_2^q, \dots, \boldsymbol{I}_n^q\}$ denotes the set of query images and $\mathcal{L}^r = \{\boldsymbol{I}_1^r,\boldsymbol{I}_2^r, \dots, \boldsymbol{I}_n^r\}$ denotes the set of reference images. We denote the 6DoF $SE(3)$ pose as 
\begin{equation}
    \boldsymbol{T} = \begin{bmatrix}
        \boldsymbol{R} & \boldsymbol{t} \\
        \boldsymbol{0}^T & 1
    \end{bmatrix}
\end{equation}
where $\boldsymbol{R} \in SO(3)$ is the rotation matrix and $\boldsymbol{t} \in \mathbb{R}^3$ is the translation in Cartesian vector space.
\subsection{System Overview} 

The structure of the proposed GNSS-Denied localization pipeline is shown in  ~\cref{system}. The VIO estimates the ego-motion of the UAV and transmits a keyframe to the VPR module at a certain frequency. The VPR module matches the query keyframe within an offline processed database where each image is cropped from the satellite map. With sufficient matches across a sequence of keyframes, the map alignment thread anchors the VIO trajectory into the Earth frame. The most recent anchored position (representing long-term memory) and the latest matching (representing instant observation) are passed to a filter to determine the UAV global position.     

\subsection{Map Alignment}
%Yao
The map alignment module aims to anchor odometry trajectory on the world map given a set of query images $\mathcal{L}^q$ with their respective positions $\{\boldsymbol{P}_i^{{L}}\}$ in odometry frame, and a corresponding set of reference images $\mathcal{L}^r$ with earth-fixed coordinates $\{\boldsymbol{P}_i^{{W}}\}$ in world frame. The reference poses are noisily corrupted, where the noise comes from:
\begin{enumerate}
    \item \textit{Cropping noise}: This noise arises from the discrete cropping of images from the satellite map, by which images are taken at certain distances. Therefore, the observed geo-references of a query image are close to the true location, but not precisely aligned. 
    \item \textit{False positive}: While our VPR provides SOTA recall in general, there can still be false positives in challenging areas, e.g., places that have repeated appearances.
\end{enumerate} 

Mathematically, the map alignment module estimates the rigid transformation $\boldsymbol{T}$ that aligns the odometry frame positions $\boldsymbol{I}_n^q$ with their corresponding noisy geo-referenced positions $\boldsymbol{I}_n^r$. 
This can be expressed as

\begin{equation}
\min_{\boldsymbol{R}, \boldsymbol{t}} \sum_{i=1}^N \| \boldsymbol{R}\boldsymbol{P}_i^L + \boldsymbol{t} - \boldsymbol{P}_i^W\|_2^2
    \label{anchor}
\end{equation}

This problem can be solved by the Iterative Closest Point (ICP) algorithm \cite{arun1987least}. However, a common issue arises due to the nature of UAVs' predominantly movement in straight lines, causing colinearity of the query and reference image positions. This colinearity results in a degenerate scenario for the ICP algorithm, leading to rotation ambiguity. 

To resolve the rotation ambiguity, we incorporate an additional constraint into the alignment problem. Specifically, we utilize gravity observability from the IMU, i.e., IMU can capture the gravity vector both in odometry and world frame. By incorporating a gravity constraint into the problem, we enforce the alignment to maintain coplanarity between the UAV's odometry trajectory and the geo-referenced map. 
\begin{equation}
\min_{\boldsymbol{R}, \boldsymbol{t}} \sum_{i=1}^N \| \boldsymbol{R}\boldsymbol{P}_i^L + \boldsymbol{t} - \boldsymbol{P}_i^W\|_2^2 + \|\boldsymbol{R}\boldsymbol{g}^L - \boldsymbol{g}^W\|_2^2
    \label{gravity constrainted}
\end{equation}

\subsection{Dealing with Drift}
To deal with long-term drift coming from odometry, we maintain a sliding window on the keyframe images, i.e., $\mathcal{L}^q$ only contains at most $N$ latest keyframes. This approach effectively reduces the impact of VIO drift over time and ensures that the query set remains relevant and up-to-date with the most pertinent keyframes. 

Lastly, we conduct an EKF to update the position, where anchored odometry provides the frame-to-frame propagation. The corrector consists of two terms:
\begin{enumerate}
    \item \textit{Long-term Pose Memory}: the latest anchored position from map alignment which captures the observation distributions within the sliding window.
    \item \textit{Instant Pose Observation}: the latest VPR observation.
\end{enumerate}

The output of EKF is the final estimated position.
% A successful map alignment relies on both reliable VIO and VPR. In the next two sections, we proceed to present our robust VIO and robust VPR module.
\subsection{Visual-Inertial Odometry}
%Yao
We use VIO to estimate the UAV ego-motion and obtain a trajectory in the odometry frame. In this process, we utilize the method proposed in \cite{he2022towards} with a monocular camera and IMU configuration. For each image, we extract Harris corner features \cite{harris1988combined} and perform feature association using Lucas-Kanade optical flow \cite{lucas1981iterative} between consecutive frames. To improve robustness, the VIO incorporates outlier rejection using 5-point RANSAC \cite{fischler1981random}. Additionally, we leverage the GPU-accelerated feature extraction and tracking provided by the NVIDIA Vision Programming Interface (VPI) \cite{vpi}.

The pipeline described in \cite{he2022towards} employs a sliding-window factor graph optimization, which jointly optimizes the reprojection residual, IMU preintegration residual, and a marginalization prior residual. During our experiments, we observed that optimization without a well-initialized seed can lead to frequent VIO failures, particularly during the fast flight of UAVs with computation constraints hardware. To address this issue, we perform an offline calibration of the IMU to obtain an initial estimation of the IMU acceleration bias. We define the IMU prior residual as described in ~\cref{BA} and incorporate it into the factor graph optimization.

% \begin{figure}[ht]
% \centering
% \includegraphics[width=0.9\linewidth]{../figures/Approach/VIO.png}
% \caption{\footnotesize The sliding window factor graph structure of the VIO, showing marginalization prior, visual, and preintegrated IMU factors. We also add an additional IMU bias prior factor to improve robustness. }
% \label{VIO}
% \end{figure}

\begin{equation}
\begin{aligned}
    	\min _{\mathcal{T}_t}\Big\{ &\|\mathbf{r}_{\mathbf{b}_a}\|^2_{\mathbf{\Sigma_{\mathbf{b}_a}}} + \|\mathbf{r}_{0}\|^2_{\mathbf{\Sigma}_0} + \\  
		&\sum_{ i \in \mathcal{L}}{\boldsymbol{\rho}(\|\boldsymbol{r}_{v_i}\|}^2_{\mathbf{\Sigma}_{{v}_i}}) 
  +\sum_{k \in \mathcal{I}}\|\boldsymbol{r}_{k_{ij}}\|^2_{\mathbf{\Sigma}_{{k}_{ij}}} \Big \} \\
\mathbf{r}_{\mathbf{b}_a} &= \mathbf{b}_a - \hat{\mathbf{b}}_a  
\end{aligned} 
 \label{BA}
\end{equation}
In ~\cref{BA}, $\mathcal{T}_t$ is the set of 6-DOF poses in the sliding window for state at current time $t$. $\mathcal{L}$ and $\mathcal{K}$ represent a set that contains visual landmarks and IMU measurements. $\boldsymbol{\rho}(\cdot)$ is the robust huber loss. $\boldsymbol{r}_{v_i}$ is the landmark reprojection residual, $\boldsymbol{r}_{k_{ij}}$ is IMU preintegration residual and $\mathbf{r}_{0}$ is the marginalization factor. Their definition can be found in \cite{qin2018vins}. 

$\mathbf{r}_{\mathbf{b}_a}$ is the IMU prior residual with $\mathbf{b}_a$ and $\hat{\mathbf{b}}_a$ as the IMU acceleration bias being estimated and pre-calibrated, respectively. Enforcing a prior knowledge of IMU bias reduces VIO failures during flight. The VIO sends keyframes to the VPR at a certain rate to obtain geo-referenced images.

\subsection{Visual Place Recognition with Foundation Model}

\label{Subsec:VPR}
Upon receiving a query image from VIO, we use a VPR module to obtain corresponding geo-referenced images from the onboard satellite imagery database. Challenges arise due to the disparity between the camera images and the satellite images. These challenges can be categorized as
\begin{itemize}
    \item \textit{Appearance variance}: The appearance of a geographical area may vary over time due to various factors such as seasonal variations, alterations in lighting conditions, changes in objects and structures, etc. 
    % \item \textbf{Different perspective}: Usually satellite images are north-aligned, while images from UAVs have arbitrary headings when UAVs revisit the same place. 
    \item \textit{Stylistic difference}: Satellite maps and UAV camera images come from distinct sensing domains, which result from varying camera parameters and rendering effects. As a consequence, the visual styles differ significantly.
    \item \textit{Recurring visual patterns}: There are areas that exhibit similar visual patterns such as forests, grassland, etc. 
\end{itemize}
~\cref{fig:nardo-air} demonstrates these challenges in our test area.

% \begin{figure}[ht]
% \centering
% \includegraphics[width=1.0\linewidth]{../figures/Approach/challenges.png}
% \caption{\footnotesize The background features a satellite-image-derived map of our area of interest. Here we highlight the distinct forest and grassland regions with recurring visual patterns. Superimposed in the center are two example onboard camera images illustrating perspective shifts and appearance variance, as well as stylistic deviations from the satellite reference image.}
% \label{challenge}
% \end{figure}

% \begin{figure}[ht]
% \centering
% \includegraphics[width=1.0\linewidth]{../figures/Approach/VPR2.png}
% \caption{\footnotesize The diagram of the VPR module. There are two main stages: (1) \textit{Offline processing} generates the VLAD centroids and database descriptors offline; (2) \textit{Online Inference} takes a query input to perform inference online and output a set of matched reference images with positions. In both of the two stages, we use the foundation model DINO ViT extractor to extract image-wise features.}
% \label{VPR}
% \end{figure}
 To address these challenges, we employ a VPR module based on the foundation model. Specifically, we utilize AnyLoc ~\cite{keetha2023anyloc} with DINO Vision Transformer (ViT)~\cite{caron2021emerging} for dense visual feature extraction. The VPR module mainly has three stages:

\subsubsection{Offline Processing}

We generate the descriptors for each image in our database at this stage. Specifically, we adopt the aerial vocabulary from AnyLoc. This vocabulary is generated using data from both VPAIR dataset \cite{schleiss2022vpair} and our Nardo-Air dataset. We choose DINO ViT extractor to obtain dense visual features from images in the aerial domain. A K-means clustering on the visual features generates $N_c$ cluster centroids, which represent the aerial vocabulary.

With the aerial vocabulary, we calculate descriptors for each satellite image in our Nardo-Air dataset. We first extract the dense features for each reference image using DINO ViT extractor. Then, we perform VLAD aggregation on these features. VLAD assigns the features to the feature vocabulary and aggregates them into image-wise VLAD descriptors. 
% During the offline processing stage, we generate the vocabulary for our satellite imagery database by leveraging the extracted features from the foundation model DINO. For a given set of $N$ satellite images, we employ the DINO ViT extractor to obtain dense visual features. These features have a shape of $C\times H \times W$, where $C$ represents the number of channels and $H$ and $W$ correspond to the down-scaled height and width of each satellite image.

% Once the dense visual features are extracted, we reshape the resulting 4-D tensor of shape $N\times C \times H \times W$ into a 2-D tensor of shape $(N \circledast H \circledast W) \times C$ to represent a global collection of pixel-wise features. Then we perform K-means clustering on the collection to obtain $N_c$ cluster centroids, which act as the vocabulary.

% Lastly, with the obtained feature vocabulary, we calculate descriptors for each image in the database. We first extract the dense features for each image with DINO. Then, we perform the VLAD aggregation on these features. VLAD assigns the features to the feature vocabulary and aggregates them into image-wise VLAD descriptors. 

\subsubsection{Online Inference}
During the online inference stage, we obtain the descriptor for each query image and retrieve the best matches in the database. Upon receiving a query keyframe from VIO, We extract dense visual features using DINO ViT extractor. Then we perform VLAD aggregation on the features using the aerial vocabulary. Lastly, we estimate the similarity between the query image and reference images to obtain the top-$K$ matches with their corresponding locations in the database.

\subsubsection{False Positive Match Filtering}
We employ a density-based clustering algorithm (DBSCAN~\cite{Ester_undated-xv}) to group the set of 2D geographic points that correspond to the retrieved matches for a given query image. We remove the points that are in single-point clusters. Then, we identify the largest cluster and return the indices of the points comprising this largest group. For localization, we then use the points from this largest cluster to calculate the reference positions $\{\boldsymbol{P}_i^{{W}}\}$.

\subsection{Selective Ordered Recall Metric}
When evaluating and comparing VPR methods a common metric used is Recall@N, which is defined as the proportion of queries for which the correct ground-truth match appears within the top-N retrievals. Additionally, some literature such as~\cite{Philbin2007-qc} and~\cite{Jegou2008-rh} also use mAP in cases where there are many correct positive matches. 

For localization purposes, where we do not employ image registration for further refinement of the database matches, we need all of the matches in the returned list of Top N matches to be close geographically, to the actual location of the query image. It is not sufficient to know whether a true-positive is somewhere in this list, but also whether there are adequate matches in the vicinity of the query image. 

Depending on the discretization and overlapping strategy used when creating the database, We consider that any given query image will capture an area that overlaps with several of our database reference images. We would then expect that the reference images with the most visual overlap with the query would have the highest similarity score, and be ranked higher in the matching sequence. 
Thus, we propose \textit{Selective Ordered Recall} metric, that takes into consideration whether a sufficient number of ground truth matches, say $k$, are present in the top $N$ of the retrieval sequence. We denote it as Top-k@N.

We formulate the metric as follows:  Let $Q$ and $R$ be the set of query images and the set of reference images in our database, respectively. For each query image $ q \in Q $, the top-$N$ closest retrieved reference images are denoted as $ \text{Retrieved}_N(q) $. The $N$ ground-truth closest reference images to $ q $ are represented as $ \text{GT}_N(q) $. 

The Selective Ordered Recall is  defined as:

\begin{equation}
\begin{aligned}
\text{Top-k@N} = \frac{1}{|Q|} \sum_{q \in Q} \delta(q)
\end{aligned}
\end{equation}

where \( \delta(q) \) is defined as:
\begin{equation}
\begin{aligned}
\delta(q) = 
\begin{cases} 
1, & \text{if } |\text{Retrieved}_N(q) \cap \text{GT}_N(q)| \geq k \\
0, & \text{otherwise}
\end{cases}
\end{aligned}
\end{equation}

In this metric, \( \delta(q) \) outputs 1 if at least $k$ out of the top $N$ retrieved reference images are in \( \text{GT}_N(q) \), and 0 otherwise. In our implementation, we care whether at least $3$ of the ground truth matches are in the top $5$ retrievals. i.e., Top-3@5. We use this metric when comparing VPR performance for our pipeline in Table \ref{tab:vpr_benchmark_test}. 

\section{Experimental Setup}
\label{Sec:setup}
\begin{figure*}[ht]
\centering
\includegraphics[width=1.0\linewidth]{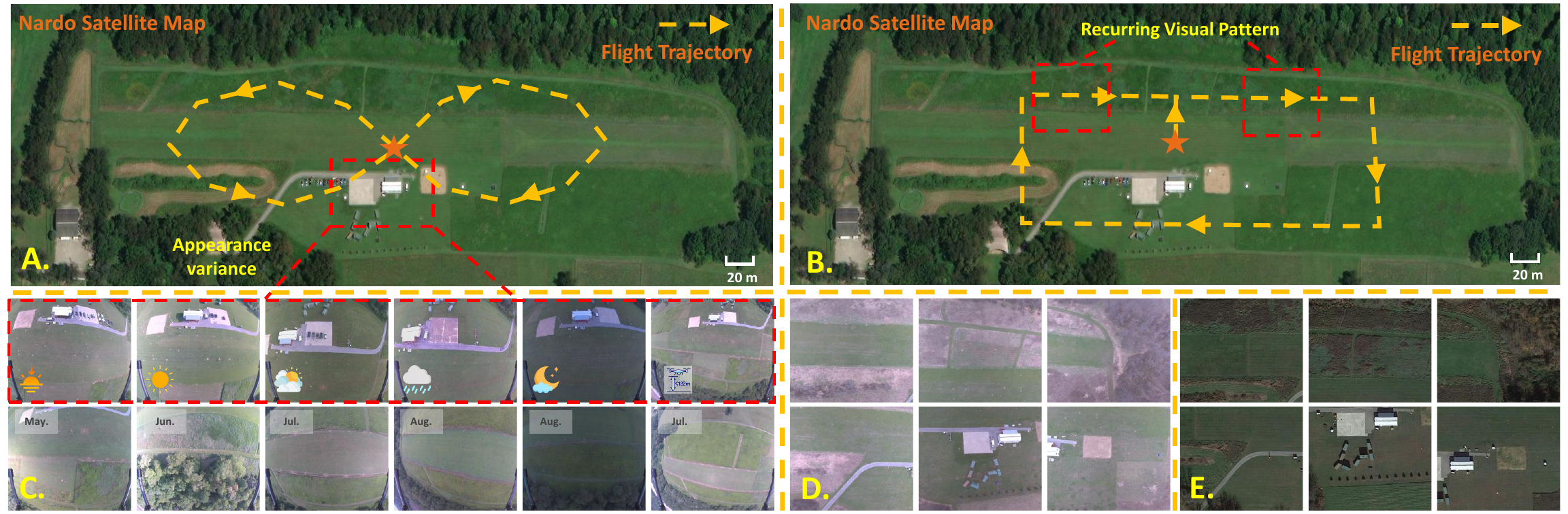}
\caption{\footnotesize\textbf{Nardo-Air} dataset. \textbf{A}: Our area of interest with one of the baseline trajectories - ``Pattern Eight''. WE highlight UAV imagery captured at the same location under different conditions. \textbf{B}: Our area of interest with one of the baseline trajectories - ``Pattern Rectangle''. We highlight an example of a recurring visual pattern. \textbf{C}: Example camera images showcasing the different seasons, lighting variations, and altitude variations in the query images of our ``Nardo-Air'' dataset. \textbf{D}: The rectified query imagery (Taken in April 2023) from our test dataset used for VPR benchmarking. \textbf{E}: The cropped reference satellite imagery (Taken in November 2021 from Google Earth) from our test dataset for VPR benchmarking and online inference.}
\label{fig:nardo-air}
\vspace{-0.5cm}
\end{figure*}

We collected the ``\textit{Nardo-Air}'' dataset, as presented in ~\cref{fig:nardo-air}, for evaluation and testing of our method. Our hardware platform is described in ~\cref{subsec:hardware_platform}. All of the data was collected at the Nardo Flight Test Field\footnote{Location: (40.591, -79.898)}. This dataset comprises UAV flight data, which includes images, IMU readings, and GPS ground truth readings. It also contains a set of reference satellite imagery used for both VPR benchmarking and database descriptor computation. We describe this dataset and its collection process in the following subsections.

\subsection{Hardware Platform}
\label{subsec:hardware_platform}
We use a custom hardware payload for both data collection and real-world flight tests. Our payload is mounted to the undercarriage of an Aurelia X6 hexacopter, in a downward-facing configuration, with the camera z-axis pointing to the ground, and the camera x-axis pointing to the left side of the hexacopter. See~\cref{fig:splash} (\textit{left}) for details.

% \begin{figure}[ht]
% \centering
% \includegraphics[width=1.0\linewidth]{../figures/experimental_setup/hardware_setup.png}
% \caption{\footnotesize Our custom hardware payload, which is mounted to the undercarriage of the Aurelia X6 hexacopter so that the camera is nadir-facing.  The sensor payload includes a camera, IMU, GPS, and an Nvidia Xavier NX for on-the-edge inference.}
% \label{hw_setup}
% \end{figure}

The specific components in our payload are:
\begin{enumerate}
  \item \textit{Camera Sensor}: Sony IMX264, which is a Type 2/3 11.1mm diagonal image sensor, with 5.07MP (2464 x 2056) effective pixel resolution, and a global shutter with a max frame-rate of $\sim$35 FPS.
  \item \textit{Camera Lens}: Commonlands CIL344, Wide-Angle 4.4mm M12 lens, with F/1.9 Resolution, with IR cut-off filter, and a $100^\circ$ Horizontal Field of View.
  \item \textit{IMU}: Epson G365, a 6 Degree of Freedom IMU with high stability, high precision, and low drift. 
  %\todo{Which parameters to include here?}
  \item \textit{GPS Module}: RadioLink SE100, a PixHawk compatible GPS receiver with up to 50cm positioning accuracy.
  \item \textit{Computer}: NVIDIA Jetson Xavier NX, with 384-core NVIDIA Volta™ GPU with 48 Tensor Cores and 6-core NVIDIA Carmel ARM®v8.2 64-bit CPU.
\end{enumerate}

In practice, due to preprocessing, temporal alignments, and other software/firmware overhead, our effective data capture metrics were the following:
\begin{enumerate}
  \item \textit{Imagery}: 24 Hz with 1224x1028 pixel resolution.
  \item \textit{IMU}: Gyroscope and Accelerometer readings at 200Hz.
  \item \textit{GPS Coordinates}: Position readings at 1Hz.
\end{enumerate}
% \todo{use a table here rather than a list?}
 
\subsection{Dataset Imagery}
The dataset imagery consists of two sets: satellite imagery and query imagery. The sampled satellite imagery comprises the database used for image matching in the VPR module.

\subsubsection{Satellite Imagery}
\label{sec:sattelite_imagery}
Our satellite imagery is sourced from Google Maps in the form of TIF imagery and is from November 2021. We deliberately exclude the infrared channel from the images, keeping the images strictly in the RGB spectrum. This imagery boasts an interpolated spatial resolution of \(0.1\, \text{m}\) per pixel. To extract relevant samples from this source TIF, a systematic sampling strategy is employed. By overlaying a grid on our designated area of interest, we ensure structured reference TIF tiling. The spacing between each sampled image tile is \(40\, \text{m}\) in all cardinal directions: North, South, East, and West. Significantly, the central point of each image sample represents the reference position for that specific database image. Therefore, if the reference image is correct for one query, the reference position error is bounded by $40/2 = 20 \text{m}$. Note that we cannot obtain altitude information in our context because satellite images usually do not provide altitude information.

Initially, every image sample extracted encompasses a dimension of \(950 \times 950\) pixels. However, to streamline our analysis and maintain uniformity, these samples undergo a resizing process. Post this adjustment, each image measured \(500 \times 500\) pixels. Examples of these reference tiles are shown in~\cref{fig:nardo-air}. D.  In terms of physical ground coverage, each cropped and resized image spans an area with a horizontal and vertical field of view (FOV) of \(60\, \text{meters}\). Given the set sampling distance and the FOV, it is noteworthy that there's approximately a \(33\%\) overlap between an image and each of its immediate neighboring images. Our area-of-interest covers approximately 142,000 $m^2$.

\subsubsection{Query Imagery}
We use the onboard camera to collect another set of imagery for VPR benchmarking. We configured a lawnmower flight trajectory that covers the entire Nardo Flight Test Field at 50m altitude. The camera is north-aligned during the flight and the camera images are extracted at 1Hz. For VPR and localization metric calculation, we performed nearest neighbors assignment based on timestamps to pair GPS position messages to corresponding images.

\subsection{UAV Trajectories \& Imagery}
% We programmed our trajectories in QGroundControl and flew the hexacopter in the waypoint-following mode for repeatability. Specifically, we mainly test two patterns: pattern eight (P.E.) and pattern rectangle (P.R.), as shown in ~\cref{}. P.E. is 880 meters long while P.R. is 1000 meters long. The UAV is flying at 10 m/s in both of the patterns. The height of flight above ground is 50 meters unless otherwise stated.
% We used the trajectory patterns described in ~\cref{tab:trajs} and ~\cref{fig:base_trajs}.

We configure the flight trajectories and conduct aerial missions with the hexacopter utilizing a waypoint-following mode for repeatability. We record the GPS data for every trajectory as ground truth. Our experiments focus on two flight patterns: ``Pattern Eight'' and ``Pattern Rectangle'', as illustrated in ~\cref{fig:nardo-air}. A and B. Pattern Eight is 880 meters long and Pattern Rectangle is 1000 meters long. The hexacopter maintains a constant velocity of $10 m/s$ throughout both patterns. Unless explicitly stated, the UAV maintained a consistent altitude of 50 meters above ground level during these missions. 

% \input{tables/table-traj}
% \begin{figure}[ht]
% \centering
% \includegraphics[width=1.0\linewidth]{../figures/experimental_setup/base_trajectories.png}
% \caption{\footnotesize Our two baseline flight trajectories in the \textbf{Nardo-Air} dataset. The area of interest is challenging and contains lots of areas with visual aliasing.}
% \label{fig:base_trajs}
% \end{figure}

% \textbf{A:} Our area of interest. We highlight a few regions with recurring visual patterns, as well as an example onboard camera image illustrating perspective shifts and appearance variance with respect to the satellite reference image (background). \textbf{B:} Our two baseline flight trajectories in the ``Nardo-Air" dataset; Pattern Eight (P.E.) and Pattern Rectangle (P.R.). \textbf{C:} A comparison of the different seasonal and temporal variations in the query images of our ``Nardo-Air" dataset. \textbf{D:} The rectified query imagery (Taken during April 2023) from our test dataset used for VPR benchmarking. \textbf{E:} The cropped reference satellite imagery (Taken at November 2021 from Google ) from our test dataset used for VPR benchmarking and online inference.
These trajectory patterns are chosen due to their conventional but challenging shape, flight distance (which we can vary by flying multiple loops of the same path), and because they are fully contained within the safe flying zone of the Nardo Flight Test Field facility. During data collection and testing flights, we keep the yaw of the vehicle fixed, i.e., always pointing towards the North, with the assumption that the compass north-aligns images for VPR. We also keep the altitude fixed, meaning that we do not localize during the ascent/descent stages.

Our UAV dataset is comprised of $20+$ trajectories with extensive flight hours from May to August 2023, covering seasonal fluctuations \SeasonChar, diverse lighting conditions \LightChar, varying altitudes \AltitudeChar (ranging from 50m to 100m), diverse flight trajectories \TrajChar, and a wide range of weather conditions \WeatherChar, as indicated in ~\cref{fig:nardo-air}. C.

\section{Results \& Discussion}
\label{Sec:result}

We conduct a series of experiments to evaluate the accuracy and robustness of our pipeline. We first present the VPR benchmarking using both the Recall@N metric and our Selective Ordered Recall metric to demonstrate our choice of VPR methods. We further present the baseline performance comparison of different variations of our methods. We then present a comprehensive analysis of \textit{FoundLoc} across In-the-Wild environmental conditions. Lastly, we demonstrate the real-time performance of \textit{FoundLoc} on our hardware. 

% We conduct a series of experiments to evaluate the accuracy and robustness of our pipeline. We first demonstrate the fundamental performance of our pipeline using the two baseline flight trajectories. Then we increase the flight altitude to demonstrate the robustness subject to altitude change, which is not addressed in previous VTRN methods. To further demonstrate the robustness of our pipeline, we collect a series of data that spans across months and various daytime encompassing diverse scenes and lighting conditions. Lastly, we present our configuration of the foundation-model-based VPR and its profile. To evaluate the accuracy of our localization results, we use the GPS signal as ground truth and evaluate the Absolute Trajectory Error (ATE) between localization and GPS position.

\subsection{VPR Benchmarking}
\begin{table}[h] % The [h] specifier suggests placing the table 'here' if possible
\centering % Centers the table on the page
\caption{VPR Performance Benchmark On Nardo-Air}
\scalebox{1.18}{
\begin{tabular}{lcccc}
\toprule
\textbf{Methods} & \textbf{R@1} & \textbf{R@5} & \textbf{Top-3@5}\\
\cmidrule{1-1} \cmidrule(lr{0.75em}){2-2} \cmidrule(lr{0.75em}){3-3} \cmidrule(lr{0.75em}){4-4}
NetVLAD & 42.25 &  76.06 & 42.25\\

NetVLAD-Fine-tuned & 78.87 & 100.0 & 52.11 \\

\rowcolor{Light}
\textbf{\textit{AnyLoc-DINO}} & 94.37 & 100.0 & 100.0 \\
 
\bottomrule
\end{tabular}
}
\label{tab:vpr_benchmark_test}
\end{table}

We evaluate both Recall@N and Top-3@5 on the classical VPR method NetVLAD and AnyLoc with Dino (AnyLoc-DINO), using our Nardo-Air Dataset. As indicated in ~\cref{tab:vpr_benchmark_test}, NetVLAD has a low R@1, R@5, and Top-3@5 without any data-specific fine-tuning. This is primarily attributed to the inherent challenges stemming from the dissimilarities between camera images and satellite images. With fine-tuning on our dataset, NetVLAD archives a significant improvement on R@1 and has 100\% R@5. However, there is a minor improvement on Top-3@5, indicating that there is a concerning amount of false positives among the top-5 retrievals even with model fine-tuning. For localization purposes, a false positive will significantly reduce the accuracy, which is intolerable. On the other hand, AnyLoc-DINO has a significantly higher Recall@N and 100\% Top-3@5 even without model fine-tuning. This guarantees reliable geo-references during the flight.
% \subsection{Unit Tests}
% In this section, we conduct unit tests on the baseline flight trajectories to demonstrate the basic performance of our pipeline. The results are shown in ~\cref{tab:Unit_test}.  While it is reasonable to expect an error of 20m because we crop each satellite image every 40m and use the center as a geo-reference position, our pipeline achieves an ATE of around 10m for the two baseline flight trajectories. This demonstrates that our pipeline is capable of estimating a promising position. 

% \input{tables/table-unit_test}

% \subsection{Tests at Different Altitudes}
\subsection{\coolname{} Performance Analysis \& Baseline Comparison}
\begin{table*}[!t]
\centering
\caption{ATE Comparison on Nardo-Air Dataset in Meters. The metrics are Average ATE (Ave.) and Standard Deviation of ATE (SD.) }
\scalebox{1.16}{
\begin{tabular}{@{}lcccccccccccccc@{}}
% \begin{tabular}{@{}clcccccccccccccc@{}}
\toprule

& \multicolumn{2}{c}{ \textbf{Season} \SeasonChar} & \multicolumn{2}{c}{ \textbf{Light} \LightChar} & \multicolumn{2}{c}{ \textbf{Altitude} \AltitudeChar} & \multicolumn{2}{c}{ \textbf{Trajectory} \TrajChar} & \multicolumn{2}{c}{ \textbf{Weather} \WeatherChar} & \multicolumn{2}{c}{ \textbf{All} \AllChar} \\ 

\cmidrule(l){2-13}

\textbf{Methods} & AVE.  & SD. & AVE. & SD. & AVE. & SD. & AVE. & SD. & AVE. & SD. & AVE. & SD. \\ 

\cmidrule{1-1} \cmidrule(lr{0.75em}){2-3} \cmidrule(lr{0.75em}){4-5} \cmidrule(lr{0.75em}){6-7} \cmidrule(lr{0.75em}){8-9} \cmidrule(lr{0.75em}){10-11} \cmidrule(lr{0.75em}){12-13}

VIO + IP & 30.92 & 31.26 & 29.26 & 24.21 & 26.26 & 30.91 & \textbf{11.32} & 1.29 & \textbf{12.07} & 3.97 & 24.18 & 22.34  \\

VIO + NetVLAD  & - & - & - & - & - & - & - & - & - & - & - & -  \\

AnyLoc-DINO  & 74.56 & 9.27 & 83.34 & 26.35 & 78.46 & 5.85 & 72.91 & 4.16 & 98.81 & 20.16 & 80.96 & 13.33 \\

\cdashmidrule{1-13}

\coolname{}{-NF} & 46.61 & 84.75 & 41.07 & 57.14 & 57.75 & 90.14 & 84.51 & 100 & 24.02 & 38.43 & 27.72 & 49.50 \\

\rowcolor{Light}
\textbf{\coolname{}} & \textbf{16.41} & 2.82 & \textbf{17.86} & 9.50 & \textbf{22.69} & 6.25 & 16.20 & 3.68 & 14.63 & 0.14 & \textbf{19.38} & 6.11 \\
\bottomrule
\end{tabular}
}
\label{tab:main-table}
\end{table*}

In this section, we analyze \coolname{}'s performance across In-the-Wild conditions and compare it with baseline methods.  Note that VTRN methods have strong assumptions on initial robot poses and use simulated odometry with known heading, so VTRN methods fail in our scenarios. We also evaluate the effect of altitude changes, which VTRN methods do not consider. The baseline methods include: VIO+IP (VIO with known initial pose, i.e., position and orientation), VIO+NetVLAD (VIO with Fine-tuned NetVLAD to provide georeference), AnyLoc-DINO (average positions of Top-3 image retrieves from AnyLoc with DINO ViT extractor), and \coolname{}-NF (\coolname{} with no false positive filtering). We use ATE as our comparison metric. The results are shown in ~\cref{tab:main-table}, with ~\cref{fig:splash} (right) plotting the ATEs.

\subsubsection{Accuracy \& Robustness}
\coolname{} achieves an average ATE below 20m at various seasons, lighting conditions, trajectory patterns, and various weather conditions, with the standard deviation of all the scenarios lower than 10m. \coolname{} maintains a consistent performance under these scenarios, demonstrating its accuracy and robustness under In-the-Wild conditions. Note that the ATE errors echo with the positional error bound in ~\cref{sec:sattelite_imagery}.
The ATE at altitude changes is slightly above 20m. This is because we do not conduct drone tilting correction. As a consequence, the keyframe sent to VPR is not exactly below the drone, causing drift when obtaining reference positions.

\subsubsection{Compared with VIO only Approach}
While VIO can achieve relative accuracy with known initial poses, it exhibits higher ATE compared to \coolname{} across various scenarios. Furthermore, the standard deviation of ATE in VIO is notably larger than that of \coolname{}. This discrepancy is primarily attributed to VIO's susceptibility to drift and scale errors, even when provided with a strong initial pose assumption. VIO tends to exhibit significant drift over the course of UAV flights. Additionally, high-altitude and high-speed flight hinder accurate scale estimation. In contrast, \coolname{} benefits from continuous drift correction through Visual Place Recognition (VPR), which effectively mitigates drift and scale errors.

\subsubsection{Compared with VPR Methods and Filtering Effect}
Even with 78.87 R@1 on the VPR benchmark dataset, NetVLAD struggles to provide reliable geo-reference positions from the satellite map. This is mainly due to the limited generalization capability of NetVLAD. In our experiments, we observe that NetVLAD keeps retrieving reference images from the center of the satellite map (framed part of ~\cref{fig:nardo-air}.A). This recurrent behavior results in failures in map alignment and, consequently, disrupts the entire pipeline.

Powered by the foundation model, AnyLoc is capable of generalizing to satellite image retrieval tasks. However, it still exhibits a large amount of false positives, decreasing the localization accuracy. This false positives effect is also reflected in \coolname{}-NF. With our false positive filtering method, \coolname{} significantly reduce the localization error.

% In this section, we vary the flight altitude while keeping the other parameters constant to evaluate the performance of our pipeline subject to altitude changes. Previous VTRN methods do not evaluate the effect of different altitudes, but altitude change is a common scenario in real flight and thus is necessary to consider.~\cref{tab:Diff_scene} demonstrates that our pipeline can maintain robust estimation for different altitudes within the range of 50m to 100m. For most of the experiments, we achieve an ATE within 20m. It is noticed that for the rectangle trajectory at 100m altitude, the ATE is abnormally large. We observe that the UAV was tilting during this flight so the places that cameras captured were not exactly below the UAV. Therefore, the matching from VPR is meters away from the UAVs' location.

% \input{tables/table-different_alt}

% \subsection{Tests with Variant Scene}
% To demonstrate the capability of dealing with variant scene appearance, we collect a series of data during different day times from May 2023 to July 2023. This dataset captures the factor of appearance variance due to seasonal variation and lighting condition change. As shown in~\cref{tab:Diff_scene}, our pipeline maintains a consistent performance under varying environmental conditions, with most of the ATE less than 20m. Benefiting from the foundation-model-based VPR, our pipeline can deal with scene variances. 
% 
% \input{tables/table-baseline_compare}

% \input{tables/table-different_condition}

\subsection{Real-time Performance of \coolname{}}
We deploy our module to onboard computational resource-limited hardware. The VIO estimates the pose at 8Hz with a total CPU consumption between $300\%$ and $400\%$ ($600\%$ maximum).  The run-time performance of the foundation model is our interest. We use the ViT-S/8 model with 21 million parameters for dense DINO feature extraction. The input images are downsampled from $500 \times 500$ pixels to $224 \times 298$ pixels. The foundation-model-based VPR achieves 1.934Hz inference frequency with above 90\% GPU usage on NVIDIA Xavier NX. VLAD encoding and matching take 0.107s and 0.0018s per query image, respectively.

% \input{tables/table-DINO_profile}

% \input{text/06-implementation.tex}

%%%%%%%%%%%%%%%%%%%%%%%%%%%%%%%%%%%%%%%%%%%%%%%%%%%%%%%%%%%%%%%%%%%%%%%%%%%%%%%%

\section{Conclusion}

\label{Sec:conclusion}
In this paper, we design and implement a vision-based GNSS-denied localization pipeline for UAVs that uses a downward-facing camera and an IMU to accurately and robustly localize in different environments. We combine a robust VIO and a robust VPR with a map alignment module to get a global estimate of our pose on a map using an onboard satellite imagery database. We utilize a foundation-model-based image matcher, that allows our system to accurately anchor with respect to \textit{appearance variances}, \textit{stylistic dissimilarities}, and \textit{recurring visual patterns} without any task or condition-specific model tuning. Our tests demonstrate that our pipeline can generate accurate localization results. We also deploy our pipeline onto an onboard system and demonstrate the possibility of running a foundation model on computation-limited hardware.

% \section*{Appendix}


\section*{Acknowledgments}

 Approved for public release; distribution is unlimited.
 This research was sponsored by DARPA (W911NF-18-2-0218).
 The views, opinions, and/or findings expressed are those of the author(s) and should not be interpreted as representing the official views or policies of the Department of Defense or the U.S. Government.
 The authors thank Jay Karhade, Avneesh Mishra, Krishna Murthy Jatavallabhula, \& Alaa Maalouf for their support with the deployment of AnyLoc. 
 We also thank John Keller for helping out with the hardware.

% Bibliography

\bibliographystyle{IEEEtran}
\bibliography{IEEEtranBST/IEEEabrv, main}

\end{document}